\documentclass[letterpaper, 10 pt, conference]{ieeeconf} 
\IEEEoverridecommandlockouts
\usepackage{url}
\usepackage{graphics} 
\usepackage{epsfig} 
\usepackage{amsmath} 
\usepackage{amssymb} 
\usepackage{color}
\usepackage{bm}
\usepackage{subcaption}
\usepackage{multirow}
\usepackage{array}
\usepackage{booktabs}
\usepackage[linesnumbered,ruled,vlined]{algorithm2e}
\usepackage{comment}
\usepackage{mathtools}
\usepackage{cite}
\usepackage{pifont}
\usepackage[capitalize]{cleveref}
\crefname{section}{Section}{Sections}
\crefname{table}{Table}{Tables}

\usepackage{cuted}
\usepackage{graphicx}
\usepackage{overpic}
\usepackage{stackengine}
\usepackage{afterpage}
\graphicspath{{figures/}}

\title{\bf MCVO: A Generic Visual Odometry for Arbitrarily Arranged Multi-Cameras }

\author{Huai Yu \quad Junhao Wang \quad Yao He \quad Wen Yang\quad  Gui-Song Xia
\thanks{Huai Yu, Junhao Wang, Wen Yang and Gui-Song Xia are with Wuhan University, Wuhan, China 430072. {\tt\small\{yuhuai, junhaowang, yangwen, guisong.xia\}@whu.edu.cn}}
\thanks{Yao He is with the Department of Electrical Engineering, Stanford University, Stanford, CA 94305, USA. {\tt\small yaohe09@stanford.edu} }
}

\begin{document}

\maketitle
\begin{abstract}
Making multi-camera visual SLAM systems easier to set up and more robust to the environment is attractive for vision robots.
Existing monocular and binocular vision SLAM systems have narrow sensing Field-of-View (FoV), resulting in degenerated accuracy and limited robustness in textureless environments. Thus multi-camera SLAM systems are gaining attention because they can provide redundancy with much wider FoV. However, the usual arbitrary placement and orientation of multiple cameras make the pose scale estimation and system updating challenging.
To address these problems, we propose a robust visual odometry system for rigidly-bundled arbitrarily-arranged multi-cameras, namely MCVO, which can achieve metric-scale state estimation with high flexibility in the cameras' arrangement. Specifically, we first design a learning-based feature tracking framework to shift the pressure of CPU processing of multiple video streams to GPU. Then we initialize the odometry system with the metric-scale poses under the rigid constraints between moving cameras. 
Finally, we fuse the features of the multi-cameras in the back-end to achieve robust pose estimation and online scale optimization. Additionally, multi-camera features help improve the loop detection for pose graph optimization. Experiments on KITTI-360 and MultiCamData datasets validate its robustness over arbitrarily arranged cameras. Compared with other stereo and multi-camera visual SLAM systems, our method obtains higher pose accuracy with better generalization ability. The code will also be made publicly available soon.

\end{abstract}

\section{Introduction}

Visual Simultaneous Localization and Mapping (vSLAM) is a fundamental technique in robotics and autonomous navigation, enabling a system to estimate its motion relative to its environment using visual sensors. Traditional monocular or stereo SLAM systems often require strict camera configurations and rely heavily on integrating inertial measurement units (IMUs) for metric-scale pose estimation and map construction \cite{1,Carlos2021Orb-slam3}.  
However, these approaches are limited by the narrow FoV and perform poorly in environments where camera placement flexibility is paramount, such as most cars with 6 cameras. 
Recently, multi-camera visual SLAM allows for greater robustness in these situations, which can cover more surrounding visual scenes to provide redundancy for poorly textured environments. It opens up new possibilities for applications such as UAV navigating through cluttered environments \cite{6}, and autonomous driving with surrounding multi-cameras \cite{8}. 
\begin{figure}
    \centering
    \includegraphics[width=0.95\linewidth]{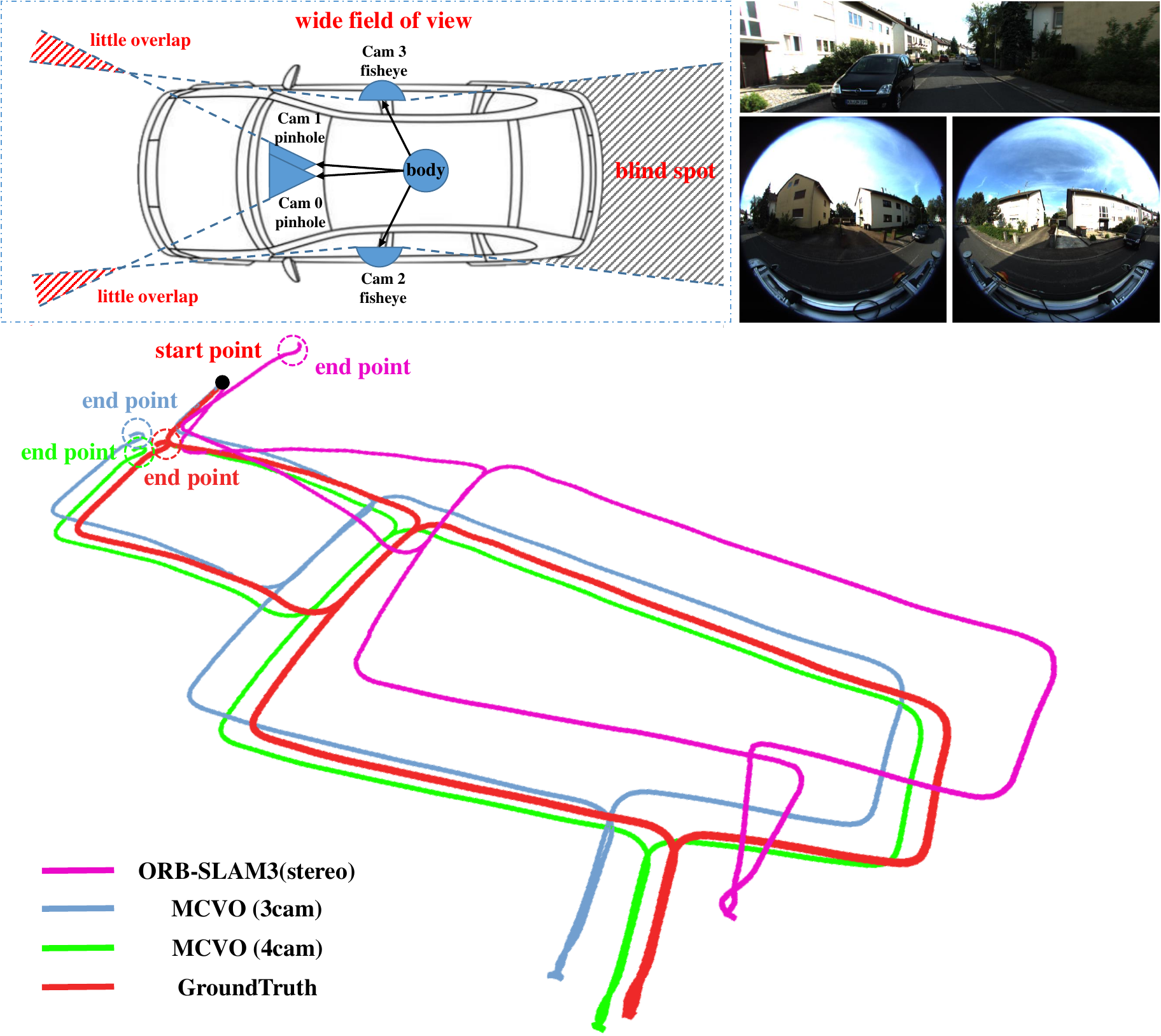}
    \caption{Illustration of the proposed MCVO system. An example of using 3-cam and 4-cam setups on KITTI-360 dataset \cite{Liao2023KITTI-360} for state estimation. The proposed MCVO performs better than ORB-SLAM3 \cite{Carlos2021Orb-slam3} using the front stereo camera.}
    \label{fig:concept}
    \vspace{-2em}
\end{figure}

However, current multi-camera visual SLAM systems still face challenges for real-world applications. On the one hand, the increase in the number of cameras, while providing information redundancy, inevitably brings several times the pressure of data processing. Most existing methods adopt conventional feature association methods, such as ORB \cite{7, 8}, which leads to a sharp increase in CPU usage and difficulty in balancing the resources for the back nonlinear optimization, resulting in the requirement of high-performance CPU or failure to achieve real-time. On the other hand, accurate scale estimation is also challenging for MCVO systems. Existing methods mostly estimate the scale by multiple binocular setups or adding IMUs \cite{10, 9341604, 20}, which require elaborate configuration of the camera's FoV overlap or extrinsic calibration between camera and IMU sensors, making it difficult to balance the flexibility of the system configuration and the accuracy of the scale estimation. Therefore, the main objective of this work is to achieve robust and generalized multi-camera visual odometry by solving the feature association computation pressure and scale estimation problem for arbitrarily arranged multi-camera systems.

To address these challenges, we propose a robust multi-camera visual odometry system, \emph{i.e.,} MCVO, which takes only multiple rigidly bundled cameras at arbitrary orientations and positions, outputting high-precision metric-scale body poses. We first design a learning-based feature extraction and tracking framework to shift the computation pressure of CPU to GPU. 
Then, we initialize the SLAM system to obtain true scale body poses based on the rigid constraint between the aligned poses for each camera using SfM. 
In the backend, we fuse multi-camera features to achieve robust pose estimation and scale optimization. The multi-camera features are further concatenated in Bag Of Words (BoW) for loop closure detection.
Through rigorous testing and practical implementation on the KITTI-360 and MultiCamData datasets, we aim to demonstrate the effectiveness of our MCVO system in enhancing the capabilities of autonomous agents across various domains with unprecedented levels of flexibility and generalizability. The highlights of the proposed system: (\emph{i}) It enables the use of multiple cameras positioned in any orientation with only the requirement of the extrinsic parameters. (\emph{ii})  By eliminating the dependency on IMU and providing the flexibility to handle cameras in arbitrary configurations, it focuses solely on visual information, no matter overlapping or non-overlapping cameras, which allows for true scale estimation and online optimization, leading to enhanced accuracy and robustness.
 (\emph{iii}) Besides, it can accommodate a variety of camera types, including but not limited to fisheye and standard pinhole cameras, making it suitable for a broad range of applications. 
The main contributions are listed as follows:
\begin{itemize}
    

    \item We propose a generic visual SLAM system framework based on universal multi-camera topology, which innovatively incorporates multi-camera feature extraction frontend, 
    metric-scale pose initialization and optimization, and multi-camera loop closure.

    \item We pioneer a robust odometry scale estimation strategy for arbitrarily arranged multi-cameras. 
    Through establishing multi-camera manifold consistency constraints on motion trajectories, we effectively calculate metric-scale poses for multi-camera systems in the initialization.
    Subsequently, we continuously refine metric scale in the backend optimization.

    \item We develop a cascaded tracking algorithm integrating SuperPoint features and Lucas-Kanade optical flow, effectively reducing computational load while enhancing loop closure through multi-camera collaborative verification.
\end{itemize}



\section{Related Work}
\label{sec:related_work}
\subsection{Efficient Feature Association Frontend}

With the widespread application of multi-camera setup in SLAM systems, several times as many images need to be processed, resulting in higher CPU usage and pressure on real-time performance. Experiments on the jetson AGX Xavier \cite{16} show that VINS-Mono consumes approximately 150-170$\%$ CPU in multi-core processing, while MSCKF exceeds 170$\%$.

To enhance system efficiency, some approaches utilize GPUs for parallel processing, shifting the computationally intensive front-end of multi-camera systems to reduce CPU usage and latency. Common feature detectors, such as Shi-Tomasi and Harris, have been implemented in CUDA Visual Library (VILIB) \cite{19} for fast feature extraction. He \emph{et al.} \cite{20} developed front-end feature tracking algorithms that run on GPUs using NVIDIA's VPI.
Additionally, learning-based methods for feature extraction and tracking \cite{21}\cite{22} have been proposed to enhance feature robustness while reducing the computational burden on the CPU. Pandey \emph{et al.} \cite{23} introduced a visual odometry algorithm combining deep learning with optical flow tracking.

To further reduce CPU usage, only selecting useful features for backend optimization can improve pose estimation accuracy while lowering the computational demands. For example, RANSAC \cite{36} is a common practice to remove obvious outliers with large reprojection errors. Scoring on the feature descriptor and uncertainty estimation using networks are also often used for selecting features for SLAM systems \cite{24, 27}.  
Carlone \emph{et al.} \cite{25} introduced an efficient feature selection algorithm using the Max-logDet metric and minimal eigenvalue, which Zhao \emph{et al.} \cite{26} later improved by optimizing the feature subset selection to maximize logDet, which makes the algorithm an order of magnitude faster than state-of-the-arts. These feature selection strategies are useful for reducing the number of features for multi-camera visual odometry systems. 

\subsection{Multiple Camera Visual Odometry}

Most current VO algorithms focus on monocular and stereo cameras, such as VINS-MONO \cite{1} and ORB-SLAM \cite{Carlos2021Orb-slam3}. While these methods offer good performance and efficiency, they are still not robust enough for autonomous pose estimation in complex real-world environments. 
Multi-camera setups can significantly increase visual constraints and provide richer image information, enhancing the robustness of SLAM systems. As a result, multi-camera SLAM has gained growing attention. Early work by Sola \emph{et al.} \cite{3} explored multi-camera VO by fusing multiple monocular camera data using filtering methods.

\begin{figure*}
\vspace{0.5em}
    \centering
    \includegraphics[trim=0 40 0 20, clip, width=1\linewidth]{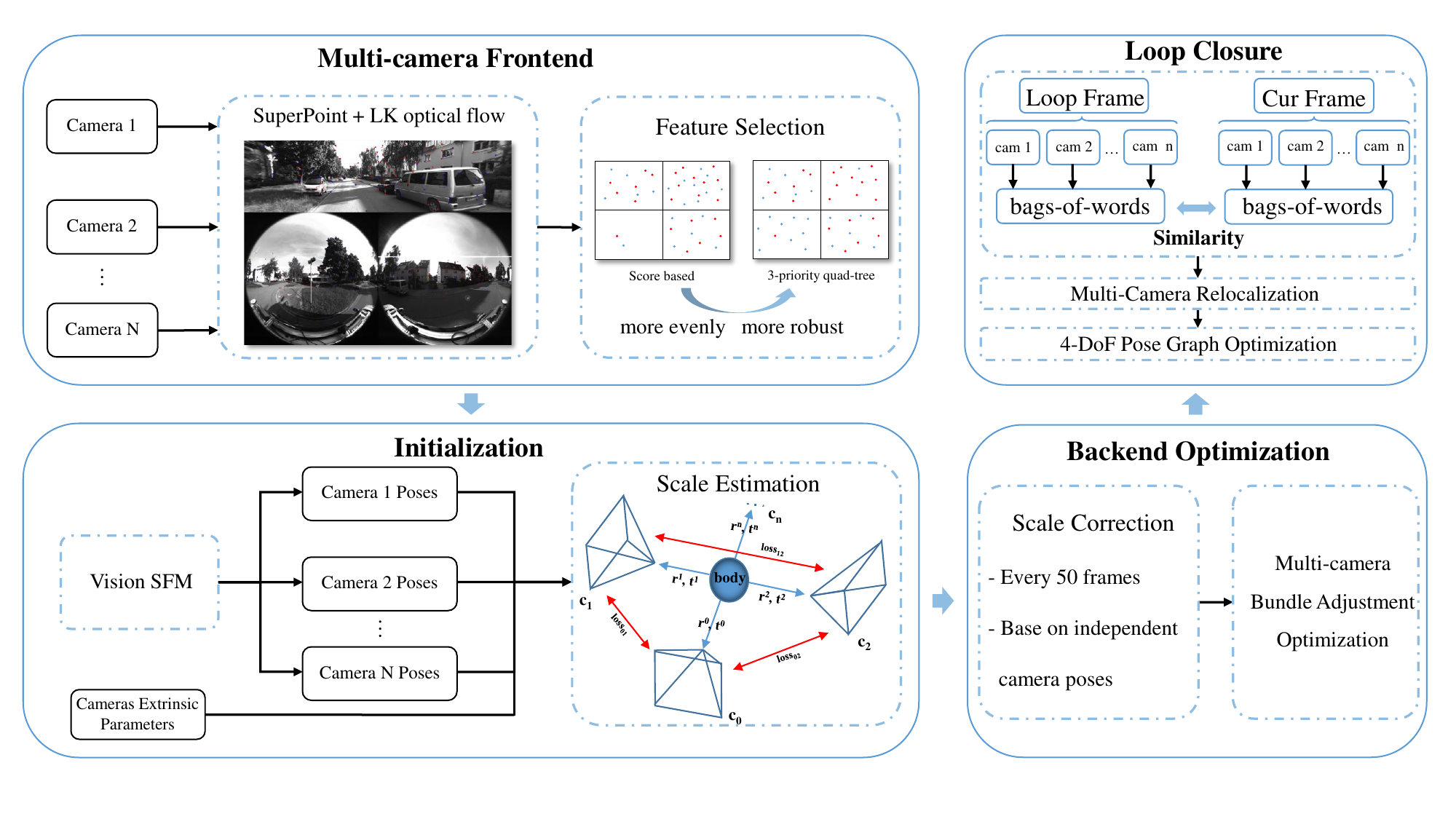}
    \caption{The pipeline of the proposed MCVO system. Scale estimation and correction ensure scale stability.}
    \label{fig:framework}
    \vspace{-1.5em}
\end{figure*}
More recent multi-camera SLAM systems \cite{6,7} extend monocular PTAM \cite{4} and ORB-SLAM to synchronized multi-camera rigs through optimization-based approaches. Liu \emph{et al.} \cite{8} proposed a pose tracker and local mapper capable of supporting arbitrary numbers of stereo cameras.
Tribou \emph{et al.} \cite{5} introduced a real-time visual odometry system utilizing multiple non-overlapping FoV cameras. This configuration provides a wider FoV, enhancing redundancy for backend optimization. 
However, it is challenging for non-overlapping cameras to obtain accurate scale information.
To improve scale accuracy, some works adopt multi-camera setups with overlapping FoV \cite{11}\cite{13}. Zhao \emph{et al.} \cite{29} proposed an early multi-camera depth estimation method, achieving stable long-term depth estimation. Kaveti \emph{et al.} \cite{10} developed a multi-camera framework that extracts 3D feature points and real scale by matching overlapping areas between cameras. Xu \emph{et al.} \cite{30} proposed a depth estimation method that does not require large FoV overlaps, enhancing depth estimation generalization across different multi-camera configurations. 
In addition to traditional scale estimation methods, many deep learning-based approaches have emerged \cite{31}\cite{32}. While these methods perform well on specific datasets, their heavy reliance on contextual information results in poor generalization in cross-dataset experiments \cite{35}. 

While overlapping multi-camera setups improve the scale estimation accuracy, they do so by reducing the range of observed information, which limits the applicability of multi-camera SLAM in non-overlapping configurations. To a certain extent, this also goes against the original intention of using multi-camera systems to maximize FoV and information capture. 
Our proposed MCVO accurately estimates scale information based on multi-camera trajectory consistency, regardless of camera overlaps. It can be applied to arbitrarily arranged camera setups, fully leveraging the wide FoV advantages of multi-camera systems.

\section{Methodology}

\label{sec:basics}
\subsection{Overview}

Our proposed multi-camera visual odometry framework is illustrated in \cref{fig:framework}. The main inputs to the framework are the synchronized multi-camera video sequences. The multiple cameras are rigidly bundled and calibrated in advance with known intrinsic and extrinsic parameters. The output is the metric scale 6-DoF robot poses in the real-world environment. The pipeline comprises four components: frontend feature extraction, pose and metric scale initialization, backend optimization, and loop closure. 
To accelerate the frontend of multi-camera feature association, we employ feature extraction with GPU acceleration and 3-priority feature selection. Then we initialize the multi-camera system with metric-scale poses using the motion invariance constraints and extrinsic parameters. To ensure the realism of the motion scale, we perform an adaptive correlation of the scale deviations during the backend optimization. Given the larger FoV of the multi-camera system, we design a more robust multi-camera omnidirectional loop detection algorithm. We further optimize the body poses in the loop with the pose graph constraint. 

\subsection{Efficient Multi-Camera Frontend}
\label{frontend}
The frontend of our proposed MCVO primarily processes multi-camera images, extracts and matches useful features, and then provides reliable landmarks for system initialization and backend optimization. 

Enhancing feature detection quality and association efficiency is critical for the visual odometry frontend. With an increasing number of cameras, the volume of extracted and tracked features grows exponentially, imposing a substantial computational burden on the edge-computing device. To address these challenges, we implement SuperPoint \cite{detone2018superpoint} for feature extraction and LK optical flow for tracking. Leveraging GPU acceleration, this framework significantly reduces CPU overhead from multi-camera setups while improving image stream processing efficiency.

To mitigate CPU load during optimization and enhance estimation accuracy, we propose a feature selection algorithm, the 3-priority quad-tree, which ensures uniform feature distribution. The conventional quadtree method clusters feature in high-texture regions, resulting in insufficient geometric constraints and compromised backend optimization accuracy. In contrast, our approach enforces spatial diversity through three criteria to mitigate feature over-concentration.
 \begin{enumerate}
     \item Quantity priority: Prioritize nodes with more features to enhance processing efficiency.
     \item Tracked feature priority: Prioritize reducing the density of points in texture-rich areas by avoiding adding useless features near tracked points.
     \item High score feature priority: Prioritize features with higher scores.
 \end{enumerate}




\begin{figure}[thpb]
		\centering
        \vspace{0.5em}
		\includegraphics[width=1\linewidth]{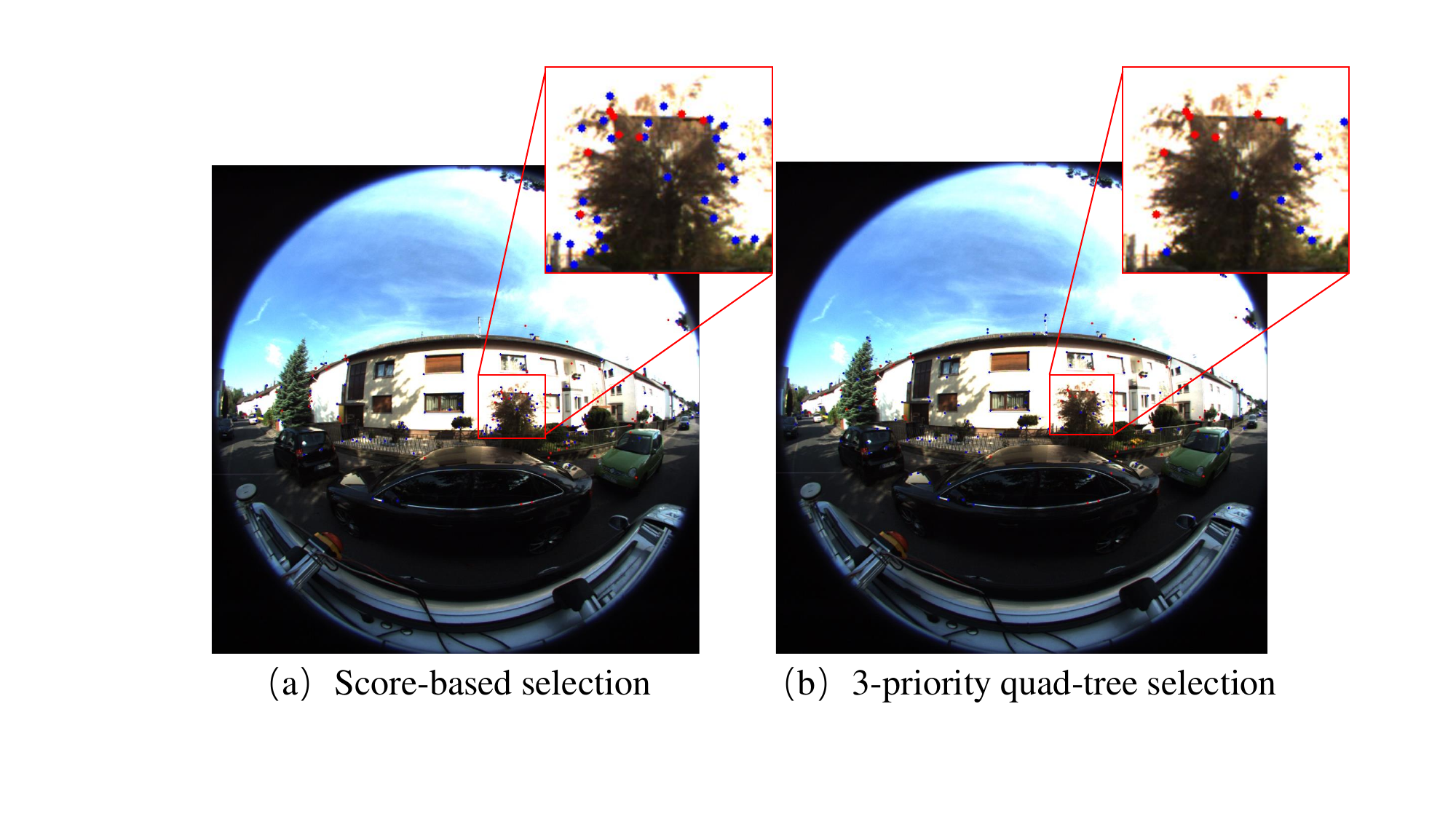}
    \caption{Distribution of feature points. The proposed approach achieves more uniform feature distribution and better robustness than the score-based method. \textit{Red points}: tracked features. \textit{Blue points}: newly added features.}
    \label{fig:featur_selection}
    \vspace{-0.4cm}
\end{figure}

\subsection{Multi-Camera Initialization}
\label{initialization}
The initialization phase of our MCVO system addresses two critical objectives: establishing initial camera poses and recovering the metric scale of motion trajectories. This process involves two components: camera pose initialization and cross-sensor metric scale calibration.

Our approach initiates by analyzing feature parallaxes within a 10-frame sliding window across all cameras. 
When the parallax of all camera streams exceeds 30 pixels at timestamp $t$, we establish the world coordinate system anchored at this timestamp. 
The camera exhibiting the highest feature tracking stability at $t$ is designated as the principal camera.
Concurrent with this spatial anchoring, we perform independent Structure-from-Motion (SfM) reconstructions for each camera using their respective observed landmarks within the initialization window. 
The pose estimations for each camera $c$ in the sliding window are mathematically represented as $\mathcal{R}^{c}$ and $\mathcal{T}^{c}$, where 
$\boldsymbol{R}^{c}_t \in \text{SO}(3)$ denotes the rotation matrix and 
$\boldsymbol{T}^{c}_t \in \mathbb{R}^{3\times 1}$ represents the translation vector at frame $t$. Then we have
\begin{equation}\label{eq:union}
\begin{aligned}
    \mathcal{R}^{c} &= \left\{ \boldsymbol{R}^{c}_t,  t \in [0, 10] \right\}, \\
    \mathcal{T}^{c} &= \left\{ \boldsymbol{T}^{c}_t,  t \in [0, 10] \right\}.
\end{aligned}
\end{equation}
It should be emphasized that these monocular SfM instances yield scale-ambiguous poses, necessitating subsequent metric scale unification through multi-camera constraints.

Concretely, building upon the camera pose estimations, we can derive the body frame poses (attached to the principal camera) through extrinsic calibration parameters $\boldsymbol{r}^{c}$ and $\boldsymbol{t}^{c}$. Notably, the scale ambiguity inherent in monocular SfM causes inter-camera scale misalignment, manifesting as inconsistent body pose estimates across sensor streams. To resolve this scale discrepancy, we introduce scale factors $\boldsymbol{s}^{c} \in \mathbb{R}^+$ for each camera to enable metric-scale body pose estimation:
\begin{equation}
\begin{aligned}
\begin{bmatrix}
    \boldsymbol{R}^{b_c}_t & \boldsymbol{T}^{b_c}_t \\
    \boldsymbol{0} & \boldsymbol{1}
    \end{bmatrix} &= \begin{bmatrix}
        \boldsymbol{R}^{c}_t & s^c\boldsymbol{T}^{c}_t \\
    \boldsymbol{0} & \boldsymbol{1} 
 \end{bmatrix}
 \begin{bmatrix}
             (\boldsymbol{r}^{c})^T & -(\boldsymbol{r}^{c})^T\boldsymbol{t}^{c} \\
    \boldsymbol{0} & \boldsymbol{1} 
 \end{bmatrix}\\
 &= 
  \begin{bmatrix}
             \boldsymbol{R}^{c}_t\boldsymbol({r}^{c})^T & -\boldsymbol{R}^{c}_t (\boldsymbol{r}^{c})^T\boldsymbol{t}^{c} + s^c\boldsymbol{T}^{c}_t\\
    \boldsymbol{0} & \boldsymbol{1} 
 \end{bmatrix}
\end{aligned}
\label{eq:body}
\end{equation}

Leveraging the theoretical motion consistency constraint that body trajectories should be scale-invariant across camera perspectives, we formulate a residual term for camera $i$ and $j$:
\begin{equation}
\begin{aligned}
 \boldsymbol{\hat{e}}(\boldsymbol{T}^{b_i}_t,\boldsymbol{T}^{b_j}_t) = &\boldsymbol{T}^{b_i}_t - \boldsymbol{T}^{b_j}_t \\
 = & -\boldsymbol{R}^{i}_t (\boldsymbol{r}^{i})^T\boldsymbol{t}^{i} + s^i\boldsymbol{T}^{i}_t + \boldsymbol{R}^{j}_t (\boldsymbol{r}^{j})^T\boldsymbol{t}^{j} - s^j\boldsymbol{T}^{j}_t \\
= & \begin{bmatrix}
    \boldsymbol{T}^{i}_t  & -\boldsymbol{T}^{j}_t
\end{bmatrix}
\begin{bmatrix}
    s^i \\ s^j
\end{bmatrix} + \boldsymbol{\theta},
\end{aligned}
\label{eq:residual}
\end{equation}
where $\theta = -\boldsymbol{R}^{i}_t (\boldsymbol{r}^{i})^T\boldsymbol{t}^{i} + \boldsymbol{R}^{j}_t (\boldsymbol{r}^{j})^T\boldsymbol{t}^{j}$. Since camera rotation matrices $\boldsymbol{R}^{c}_t$ are scale-independent, $\boldsymbol{\theta}$ is related to the camera extrinsic parameters $[\boldsymbol{r}^c, \boldsymbol{t}^c]$ and camera rotation matrices $\boldsymbol{R}^{c}_t$, thus $\boldsymbol{\theta}$ will be also scale-independent. Building upon the pairwise residual formulation in \eqref{eq:residual}, we extend the constraint to $N$-camera systems through block matrix construction. Let the scale vector encapsulate all camera scale factors, the residual takes the form:
\begin{equation}
\begin{aligned}
  \boldsymbol{e}(\boldsymbol{T}^{b_i}_t, \boldsymbol{T}^{b_j}_t) 
&= 
\begin{bmatrix}
    \boldsymbol{0}_\alpha & 
    \boldsymbol{T}^{i}_t & 
    \boldsymbol{0}_\beta & 
    -\boldsymbol{T}^{j}_t &  
    \boldsymbol{0}_\gamma
\end{bmatrix}
 \boldsymbol{s} + \boldsymbol{\theta} \\
&= \boldsymbol{F}\boldsymbol{s} + \boldsymbol{\theta}, \\
\boldsymbol{s} &= \begin{bmatrix}s^1 & s^2 & \ldots & s^N \end{bmatrix}^T
\end{aligned}
\label{eq:multi_residual}
\end{equation}
where $\alpha=3\times(i - 1)$, $\beta=3\times(j-i-1)$, $\gamma=3\times(N-j), (i<j)$. $\boldsymbol{F} \in \mathbb{R}^{3\times N}$ forms the sparse observation matrix. This construction preserves the problem's sparsity pattern for efficient large-scale optimization. 
$\boldsymbol{\theta}$ denotes the $\boldsymbol{s}$-independent component in Equation \eqref{eq:residual}.
The complete scale estimation problem then reduces to solving the constrained least-squares minimization:
\begin{equation}
\begin{aligned}
    \min_{s_1, s_2,..., s_{N}} \sum_{t}\sum_{i=1}^{N}\sum_{j=i+1}^{N} &|| \boldsymbol{e}(\boldsymbol{T}^{b_i}_t,\boldsymbol{T}^{b_j}_t) ||^2
\end{aligned}
\label{SO}
\end{equation}
The scale estimation is formulated as a manifold-aware optimization problem within the ceres solver \cite{agarwal2019ceres}, with scale factor $\boldsymbol{s}$  parameterized over Lie groups. We solve the optimization problem using the Levenberg-Marquardt algorithm, which effectively handles the sparse Jacobian structure, with automatic differentiation ensuring numerical precision in partial derivative calculations.

\subsection{Scale Correction and Backend Optimization}
\label{backend}
Following metric scale initialization, we formulate a multi-camera visual bundle adjustment problem that jointly optimizes camera poses and 3D landmarks by minimizing the Mahalanobis-weighted reprojection residuals:
\begin{equation}
\begin{aligned}
    	\min _{\mathcal{X}} & \Big\{\|\boldsymbol{r}_{0}\|^2_{\mathbf{\Sigma}_0} + \sum_{c\in N}\sum_{f\in\mathcal{F}_c}{\boldsymbol{\rho}(\|\boldsymbol{r}_{v_{(c,f)}}\|}^2_{\mathbf{\Sigma}_{{v}_{(c,f)}}}) \Big \},  \\
    \mathcal{X} &= [\boldsymbol{x}_{1}^b, \boldsymbol{x}_{2}^b, ..., \boldsymbol{x}_{t}^b, \boldsymbol{\lambda}_1, \boldsymbol{\lambda}_2, \cdots, \boldsymbol{\lambda}_{N} ],
\end{aligned}
\label{BA}
\end{equation}
where $\boldsymbol{x}_{t}^b$ denotes the body poses in the sliding window and  $\boldsymbol{\lambda}_c$ denotes inverse-depth parameterized landmarks observed in camera $c$.  $\mathcal{F}_i$ is the set of reliable features in $i$-th camera, $\boldsymbol{r}_{0}$ is the marginalization factor, $\boldsymbol{r}_{v_{(i,j)}}$ is the $j$-th landmark reprojection residual in the $i$-th camera. 
$\boldsymbol{\rho}(\cdot)$ implements Huber robustification. 
The Ceres solver's Dogleg algorithm \cite{agarwal2019ceres} efficiently handles this nonlinear least-squares problem through adaptive trust region management and analytical Jacobians derived from \cite{1}.

\textbf{Scale Correction}.  While the iterative bundle adjustment process continuously refines body poses and feature depths through nonlinear optimization, inherent scale ambiguity leads to progressive scale drift during map expansion. To mitigate this critical issue, we implement a hierarchical scale correction module that operates during the backend optimization.  Subsequently, we compute the optimal scale factors $s$ through the constrained optimization framework detailed in \cref{initialization}. 
Rather than applying direct scale transformation to pose trajectories, we optimally adjust inverse feature depths based on scale factors within each camera's coordinate system. 
This differential depth propagation approach enables implicit scale rectification through subsequent state estimation updates while maintaining the temporal consistency of the SLAM system.

\subsection{Multi-Camera Loop Closure}
\label{loop_closure}

Despite high-accuracy multi-camera VO and scale correction, drift accumulation persists. To address the problem, we design a cross-camera loop closure module.

\begin{figure}[thpb]
      \centering
      \includegraphics[width=1\linewidth]{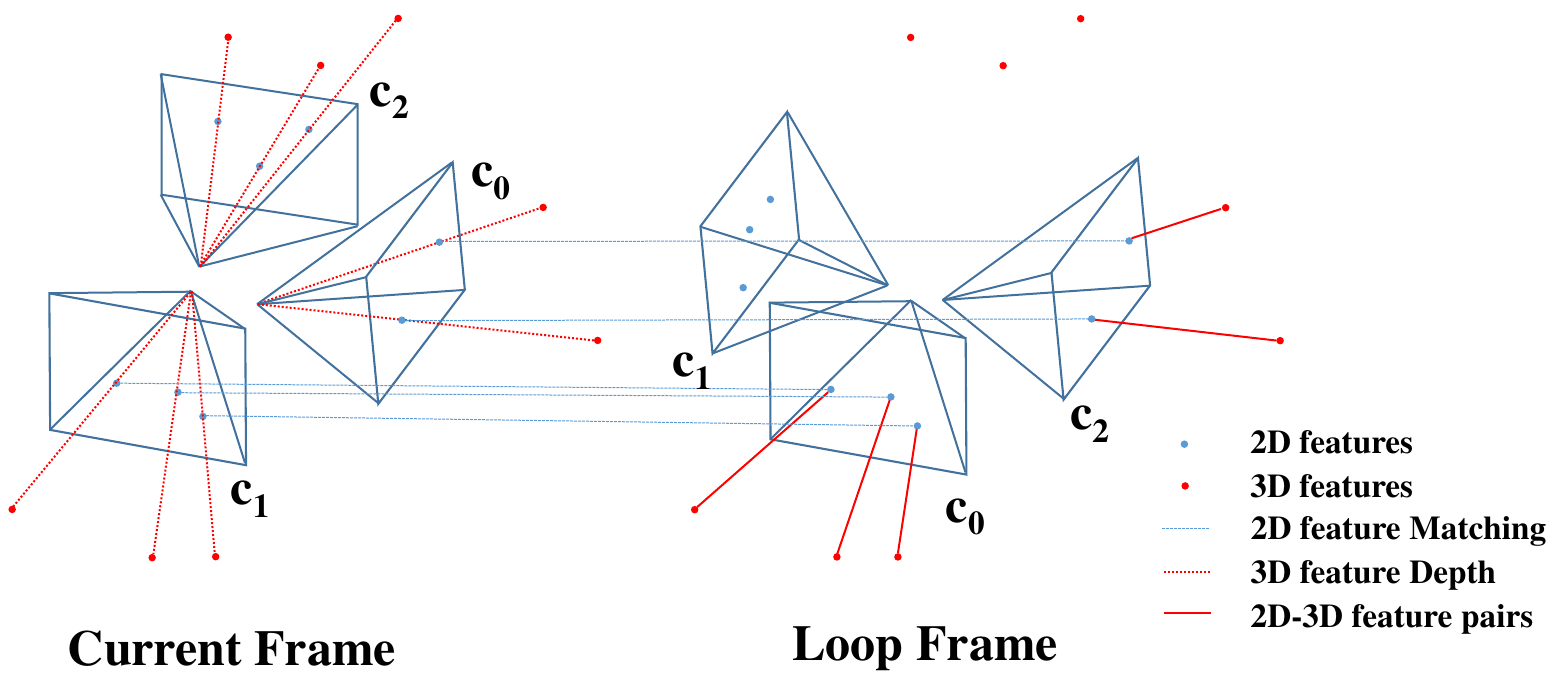}
      \caption{Illustration of the multi-camera re-localization}
      \label{fig:mc_loop}
   \end{figure}

\textbf{Loop Detection}. We construct a cross-camera feature fusion mechanism based on the DBow2\cite{GalvezTRO12} model. By integrating synchronously captured features from multi-camera into a unified feature space and enforcing spatiotemporal consistency constraints for inter-frame feature matching, we establish cross-camera shared bag-of-words representations. Specifically, as shown in Fig. \ref{fig:mc_loop}. This methodology effectively mitigates geometric constraint degradation caused by viewpoint disparities, significantly enhancing the robustness of loop detection in complex multi-camera environments. 


\begin{table*}[ht]
\vspace{0.5em}
\belowrulesep=0pt
\aboverulesep=0pt
\centering
\caption{\small Comparison of ATE on various datasets. The metrics translation, rotation, and scale are in meters, degrees, and percentages, respectively. Note that MultiCamSLAM fails after running a certain distance (as shown in parentheses) on the KITTI360.} 
\label{tab:ATEtable}
\resizebox{\textwidth}{!}{
\renewcommand{\arraystretch}{1.5}
\begin{tabular}{c|ccc|ccc|ccc|ccc|cc|cc}
\toprule
\multirow{3}{*}{Method} & \multicolumn{3}{c|}{KITTI360\_00} & \multicolumn{3}{c|}{KITTI360\_03} & \multicolumn{3}{c|}{KITTI360\_05} & \multicolumn{3}{c|}{KITTI360\_10} & \multicolumn{2}{c|}{MCdata\_Lab1} & \multicolumn{2}{c}{MCdata\_Ground1} \\ [-1ex]
& \multicolumn{3}{c|}{(1520m)} & \multicolumn{3}{c|}{(1379m)} & \multicolumn{3}{c|}{(1173m)} & \multicolumn{3}{c|}{(3343m)} & \multicolumn{2}{c|}{(152m)} & \multicolumn{2}{c}{(90m)} \\
[-0.8ex] & Trans. & Rot. & Scal. & Trans. & Rot.& Scal. & Trans. & Rot. & Scal. & Trans. & Rot. & Scal. & Trans. & Trans. & Trans. & Trans.\\
\midrule
VINS-Fusion & 76.188 & 30.918 & 1.008 & 38.213 & 37.502 & 0.556 & 14.516 & 30.054 & 0.181 & 39.235 & 33.082 & 1.464 & 16.579  & 10.921\% & 14.80  & 16.469\% \\
ORB-SLAM3 & 22.433 & 11.262 & 0.396 & 12.744 & 6.555 & 0.346 & 8.320 & 5.868 & 0.171 & 34.842 & 8.477 &1.336 & \textbf{2.90}  & \textbf{1.91\%} & \textbf{0.85} & \textbf{0.80\%}\\
MultiCamSLAM & 15.307(245.2) & 16.712 & 0.330 & 8.716(222.5) & 10.382 & 0.181 & 6.744(91.9) & 55.356 & 0.119 & 56.413(553.5) & 43.253 & 1.568  & 14.83  & 9.75\% & 1.69 & 1.88\% \\
MCVO(ORB) & 8.613 & \underline{8.241} & 0.134 & 12.690  & 4.542 & 0.368  & 11.986  & 3.646 &  0.145  &  36.316 &  5.812 &  1.503 & 11.50 & 7.561\% & 14.58 & 16.321\% \\
MCVO(2-cam) & \textbf{6.059} & 8.459 & \textbf{0.105} & \underline{11.829} & 5.179 & \underline{0.272} & \underline{6.215} & 7.274 & \underline{0.062} &\underline{24.826} & 7.947 & \underline{1.007} & 5.85 & 3.85\% & 4.35 & 4.83\% \\
MCVO(3-cam) & 13.933 & \textbf{6.847} & 0.300 & 20.234 & \underline{2.585} & 0.394 & 6.270 & \textbf{2.578} & 0.065 & 36.599 & \textbf{4.212} &1.262 & 7.89 & 5.20\% & 1.54 & 1.711\% \\
MCVO(4-cam) & \underline{7.610} & 8.303 & \underline{0.129} & \textbf{4.889} & \textbf{2.256} & \textbf{0.138} &\textbf{3.469} &\underline{3.264} &\textbf{0.037} &\textbf{22.262} & \underline{5.418}  & \textbf{0.876} &\underline{5.50} & \underline{3.62}\% &\underline{0.95} & \underline{1.06}\%\\
\bottomrule
\end{tabular}
}
\end{table*}

\begin{figure*}[htbp]
	\centering
	\begin{subfigure}{0.49\linewidth}
		\centering
		\includegraphics[width=1\linewidth]{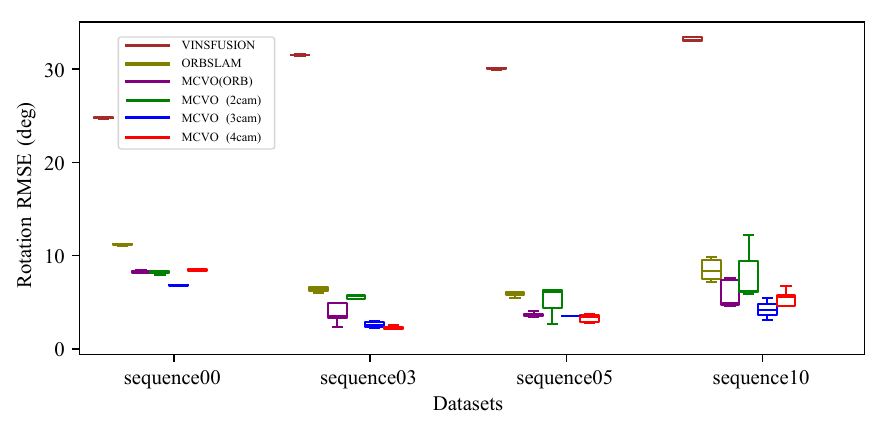}
		\label{fig:ATE_trans}
	\end{subfigure}
	\centering
	\begin{subfigure}{0.49\linewidth}
		\centering
        \includegraphics[width=1\linewidth]{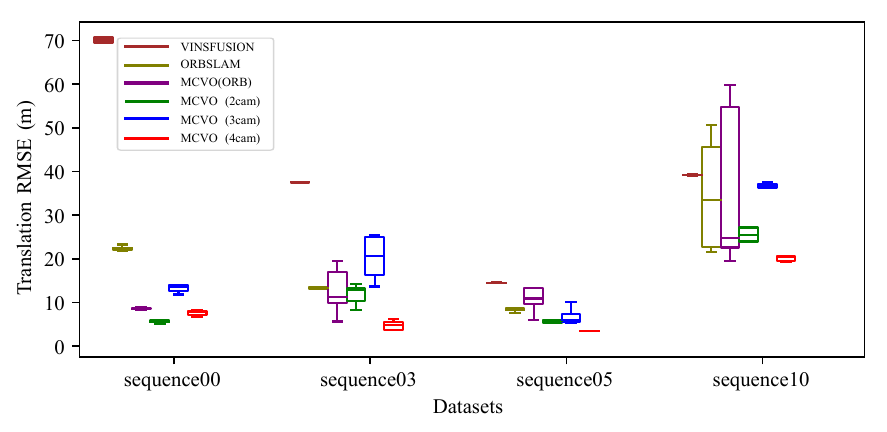}
		\label{fig:ATE_rot}
	\end{subfigure}
    \vspace{-0.5cm}
    \caption{Comparison of ATE for different methods on the KITTI360}
    \label{fig:ATE}
    \vspace{-0.5cm}
\end{figure*}

\textbf{Pose Graph Optimization} We employ 6-DoF pose graph $\boldsymbol{G} = \{\boldsymbol{V}, \boldsymbol{E}\}$ optimization \cite{1} to refine trajectories using vertices $\boldsymbol{V}$ (keyframes) and edges $\boldsymbol{E}$ (transformations). Marginalized keyframes from the sliding window are incorporated into $\boldsymbol{G}$ with relative transformations as edges.   Upon loop detection, a loop closure edge triggers 6-DoF optimization, mitigating accumulated errors and enhancing robustness through spatiotemporal constraint propagation.


\section{Experiments}

\label{sec:experiments}
\subsection{Experimental Setup}
\subsubsection{Datasets}
For the evaluation of our method, we require datasets containing multi-camera setups with pre-calibrated intrinsic and extrinsic parameters.
We select two publicly available benchmarks: \emph{KITTI-360} \cite{Liao2023KITTI-360} and \emph{MultiCamData} \cite{10}, which exhibit complementary characteristics in terms of camera configurations, environmental diversity, and operational challenges.
The outdoor-oriented KITTI-360 dataset comprises 4 cameras with heterogeneous configurations: two $180^{\circ}$ fisheye and two $90^{\circ}$ pinhole cameras arranged on a vehicle (Fig. \ref{fig:concept}). This dataset provides large-scale urban road scenarios 
with three principal challenges: (1) Illumination variations, (2) Texture degradation in wilderness, and (3) Dynamic objects. We evaluated on 4 sequences for rigorous trajectory comparison with ground truth.
Besides, the indoor MultiCamData features 6 pinhole cameras comprising four forward and two side cameras. It presents distinctive challenges including (1) Feature scarcity in narrow corridors, and (2) Severe texture degradation on blank walls. The highly constrained environments enable effective verification of the robustness and generalization ability of the proposed method.


\subsubsection{Evaluation Setup} We evaluate against three state-of-the-art visual odometry systems: VINS-Fusion \cite{1}, ORB-SLAM3 \cite{Carlos2021Orb-slam3}, and MultiCamSLAM \cite{10}. For metric consistency, VINS-Fusion and ORB-SLAM3 utilize calibrated stereo configurations, while MultiCamSLAM follows its native multi-camera implementation[10]. Our method adopts four-camera fusion on KITTI-360: two fisheye + two pinhole cameras, with ablative configurations (2-cam: dual pinhole; 3-cam: dual fisheye + mono pinhole).
Evaluation metrics comprise Absolute Trajectory Error (ATE), Relative Pose Error (RPE), and scale drift\cite{Zhang18iros} for KITTI-360, while MultiCamData evaluate metrics via visual marker-based start-end pose discrepancy \cite{10} due to the absence of ground truth trajectories. All experiments are conducted on a desktop with an Intel Core i7-11700 CPU and an NVIDIA RTX3060 GPU.

\subsection{Experimental Results}
The quantitative results for the ATE metric are shown in \cref{tab:ATEtable} and \cref{fig:ATE}. Stereo VINS-Fusion shows much worse results than stereo ORB-SLAM3. 
Compared with the stereo ORB-SLAM3 and VINS-Fusion, MCVO achieves lower rotation and translation error on most of the sequences of KITTI360.
This improvement stems from two key advantages: (1) the robust scale estimation module reduces scale drift across all sequences, and (2) the multi-camera configuration enables stable feature tracking through wide FoV coverage.
MultiCamSLAM fails to complete most sequences in KITTI360 due to inadequate handling of large inter-frame displacements in high-speed scenarios, preventing reliable triangulation and initialization.

Experimental results of MCVO with varying camera configurations are presented in the last three rows of  \cref{tab:ATEtable}. We first compare MCVO under 4 camera setup using ORB feature and the proposed learning-based front-end. The comparison shows the robustness of our front-end design. In terms of translation and scale estimation, 3-cam setup operates in a fully non-overlapping configuration, significantly outperforming ORB-SLAM3 and VINS-Fusion with stereo setup, which can be attributed to our robust multi-camera scale estimation strategy. Both 2-cam and 4-cam setups employ forward stereo cameras, where their overlapping FoV contributes to enhanced translation accuracy. Notably, the 4-cam setup generally achieves superior performance compared to the 2-cam setup, demonstrating the effectiveness of multi-camera configurations in improving state estimation accuracy.
Regarding rotation estimation, multi-camera configurations significantly outperform stereo setups, as their distributed FoV maintains feature observability during rapid rotations, extending tracking continuity by 3-5 frames.

Moreover, MCVO demonstrates consistent robustness on MultiCamData, notably outperforming VINS-Fusion in pose estimation accuracy while being competitive to ORB-SLAM3. 
This minor precision gap stems from texture-deprived environments like uniform white walls, that degrade feature matching fidelity. 
Compared to MultiCamSLAM, which also employs a non-overlapping camera setup, MCVO achieves superior accuracy and better generalization across different sequences.

\begin{figure}[h]
    \centering
    \begin{subfigure}{1\linewidth}
        \centering
        \includegraphics[width=0.95\linewidth]{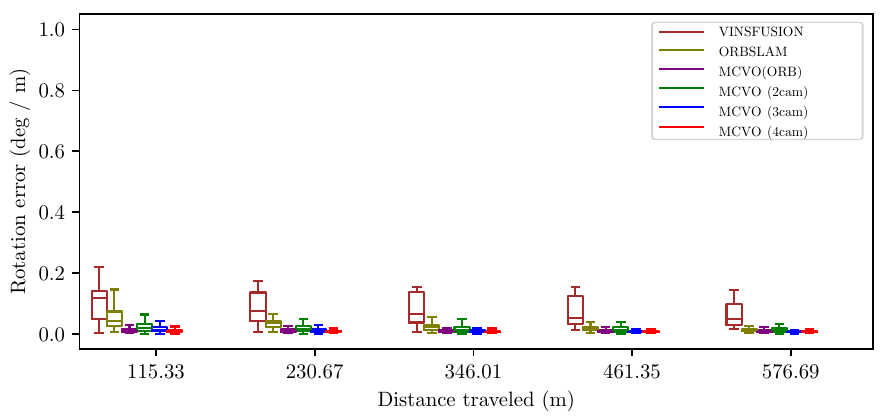}
    \end{subfigure}\\
    \begin{subfigure}{1\linewidth}
        \centering
        \includegraphics[width=0.95\linewidth]{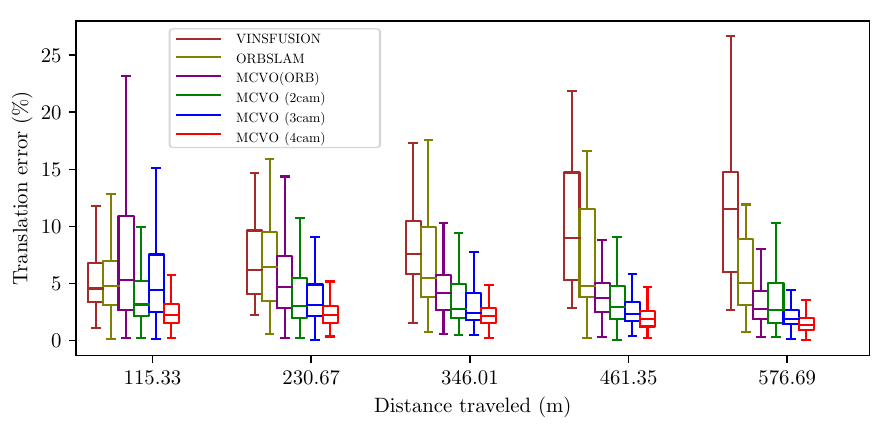}
    \end{subfigure}
    \caption{Comparison of RPE for different methods on the KITTI360\_05 sequence.}
    \vspace{-2em}
    \label{fig:RPE}
\end{figure}
\begin{figure}[h]
\vspace{0.5em}
    \centering
    \begin{subfigure}{1\linewidth}
        \centering
        \hspace{-0.5cm}
        \includegraphics[width=0.85\linewidth]{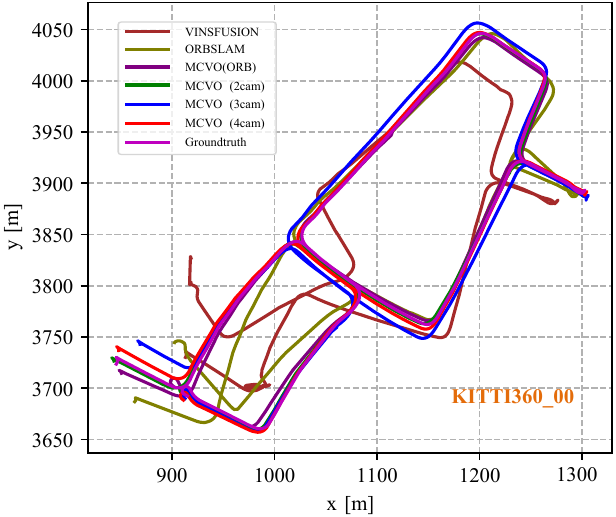}
        \label{fig:KITTI360_00}
    \end{subfigure}\\
    \begin{subfigure}{1\linewidth}
        \centering
        \hspace{-0.65 cm}
        \includegraphics[width=0.9\linewidth]{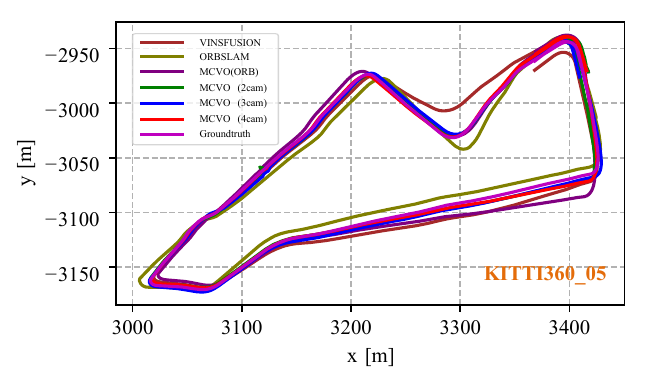}
        \label{fig:KITTI360_05}
    \end{subfigure}
    \caption{Trajectories of different sequences on  KITTI360.}
    \label{fig:traj_KITTI360}
    \vspace{-0.3cm}
\end{figure}

Fig. \ref{fig:RPE} presents the RPE results of different methods on the 05 sequence of KITTI360. MCVO (4-cam) outperforms other methods with smallest pose drifts. This improvement is largely due to our continuous correction strategy and multi-camera wide FoV. These results validate the effectiveness of MCVO on state estimation. 
To further qualitatively analyze the performance, we plot the trajectories of different methods on sequences 00 and 05 of KITTI360 in \cref{fig:traj_KITTI360}. ORB-SLAM3 and VINS-Fusion exhibit large rotation errors on the 00 sequence. In contrast, MCVO shows improved stability and accuracy under the same conditions. 

\subsection{Ablation Study}

In this subsection, we conduct ablation study on the front-end and back-end of MCVO to verify their effectiveness.

\begin{table}[ht]
\belowrulesep=0pt
\aboverulesep=0pt
\centering
\caption{MCVO performance under different feature extractors on the KITTI360 00 sequence.}
\label{tab:feature}
\scalebox{0.7}{
\renewcommand{\arraystretch}{1.2}
\begin{tabular}{c|cc|cc|c|c|c}
\toprule
\multirow{2}{*}{Method} & \multicolumn{2}{c|}{Transl.[m]} & \multicolumn{2}{c|}{Rot.[°]} & \multirow{2}{*}{Usage$_{CPU}$[\%]}  &\multirow{2}{*}{Usage$_{GPU}$[\%]}  & \multirow{2}{*}{FPS} \\
& Mean & Std & Mean & Std & & & \\
\midrule
Shi-Tomasi(CPU) &9.413 & 9.448 & 8.570 & 8.559 & 534.68 & - & 18.75\\
ORB(CPU) & 8.613 & 8.595 & \bf{8.241} & \bf{8.217}  & 942.4 & -  & 12.74\\
Superpoint(CPU) &28.841 & 26.653 & 9.553 &9.108 & 1023.6 & - & 9.02\\
SuperPoint(GPU) & \bf{7.610} & \bf{7.793} & 8.303 & 8.445 & \bf{495.42} & 43.75
   & \bf{19.50}\\
\bottomrule
\end{tabular}
}
\vspace{-0.3cm}
\end{table}

\begin{table}[ht]
\vspace{0.5em}
\belowrulesep=0pt
\aboverulesep=0pt
\centering
\caption{{MCVO performance under different feature matching methods (Superglue-SG).}}
\label{tab:match}
\resizebox{0.5\textwidth}{!}{
\renewcommand{\arraystretch}{1.5}
\begin{tabular}{c|cc|c|c|c}
\toprule
\multirow{2}{*}{Method} & \multicolumn{2}{c|}{Inlier} & {Usage$_{CPU}$} & Aver. track& {Initialization }\\
& Num & ratio & Mean  & times & succ.\\
\midrule
ORB+KNN & 51.33 & 34.22\%  & 892.03\%  & 2.71 & \ding{53}  \\
ORB+LK & 93.04  & 62.03\%  & 750.78\%  & 15.95 & \ding{51}  \\  
\hline
SuperPoint+SG & 65.23 & 43.39\% & 789.53\% & 4.32 & \ding{53} \\ 
SuperPoint+LK & \bf{99.61} & \bf{66.41\%} & \bf{710.12\%} & \bf{16.40} & \ding{51} \\
\bottomrule
\end{tabular}
}
\end{table}

\begin{table}[ht]
\belowrulesep=0pt
\aboverulesep=0pt
\centering
\caption{MCVO performance under different feature selection on the KITTI360 00 sequence. (Spatial Feature Distribution-SFD)}

\label{tab:feature_selection}
\scalebox{1}{
\renewcommand{\arraystretch}{1.2}
\begin{tabular}{c|c|cc|cc}
\toprule
\multirow{2}{*}{Method} &\multirow{2}{*}{SFD $\downarrow$} & \multicolumn{2}{c|}{Transl.[m]} & \multicolumn{2}{c}{Rot.[°]}  \\
& &    Mean    &    Std    &    Mean    &    Std     \\
\midrule
score-based & 24.25  & 8.740 & 8.678 & 8.617 & 8.515  \\
quad-tree &22.18 &  8.432& 8.437& 8.680 & 8.633   \\
3-priority & \bf{21.95}  & \bf{7.610} & \bf{7.793} & \bf{8.303} & \bf{8.445} \\
\bottomrule
\end{tabular}
}
\vspace{-0.3cm}
\end{table}

\subsubsection{Feature extractor}
Experimental evaluations comparing feature extractors across pose accuracy and computational efficiency metrics reveal SuperPoint's superior performance (\cref{tab:feature}). It achieves a 18.4\% higher pose accuracy than conventional Shi-Tomasi and ORB, attributable to its multi-scale feature extraction capability and enhanced robustness to environmental variations. Furthermore, MCVO (SuperPoint) reduces CPU utilization by 47.4\% compared to ORB, enabled by the GPU-accelerated frontend architecture. This efficiency gain facilitates resource allocation for backend optimization processes, ensuring both high-precision pose estimation and real-time operational efficiency.


\subsubsection{Feature matching}
As shown in \cref{tab:match}, feature matching based on LK flow in MCVO demonstrates superior computational efficiency and robustness  compared to descriptor approaches. Quantitative analysis of feature tracking duration reveals that the LK flow maintains the longest feature lifespan (avg. 16.18 frames), providing stable landmark constraints for backend optimization. In contrast, other methods exhibit short tracking lifespans, resulting in a high failure rate during pose initialization.


\subsubsection{Feature selection}
As shown in \cref{tab:feature_selection}, we conducted ablation studies on the feature selection algorithm. We introduce a new metric, SFD, to quantify Spatial Feature Distribution by computing the feature density on uniform grids of images and derive the variance across all cells. Compared with score-based methods, our 3-priority algorithm demonstrates 9.48\% higher feature dispersion, effectively alleviating the issue of excessive feature concentration in texture-rich regions. The proposed method also achieves 11.0\% improvement in pose estimation accuracy over both quadtree and score-based algorithms, validating its effectiveness.
\subsubsection{Loop closure}
\begin{figure}[thpb]
  \centering
  \includegraphics[width=0.85\linewidth]{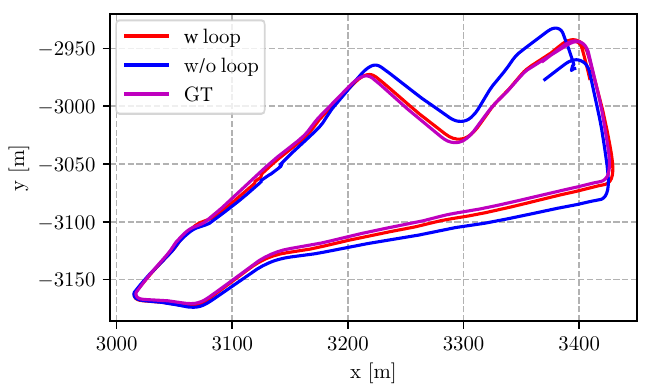}
  \caption{Loop closure effectiveness validation on the KITTI360 05 sequence.}
  \label{fig:loop}
  \vspace{-0.5cm}
\end{figure}

To validate the efficacy of the multi-camera loop closure, we conducted systematic evaluations on the KITTI360 05 sequence. Qualitative analysis (\cref{fig:loop}) demonstrates that our multi-camera closure framework enables robust loop detection and closure, establishing reliable global constraints for backend optimization. The proposed MCVO effectively mitigates cumulative mapping errors through this mechanism, achieving enhanced state estimation accuracy.

\subsection{Discussion}
The comparison with other stereo VO and multi-camera VO demonstrates the effectiveness of the proposed MCVO on state estimation. The analyses of each component validate the reasonableness of our design on multi-camera data processing. However, there are still limitations:
\begin{itemize}
    \item Multi-camera configurations can improve VO robustness, but inevitably introduce new computational requirements. Our approach uses learning-based feature extraction to shift CPU consumption but introduces the need of GPU resources. The best approach may be to develop a specialized camera with feature association embedded in the sensor hardware.
    \item MCVO requires the extrinsic parameters of arbitrarily arranged multi-cameras. When there is no FoV overlap between cameras, precise extrinsic calibration will also be a challenging problem.
\end{itemize}

\section{Conclusions and Future Work}
\label{sec:conclusion}
In this paper, we propose a robust multi-camera visual odometry system, which has high flexibility on the camera setup with different camera types, with or without FoV overlap. The arbitrary arrangement of multi-cameras improves the flexibility of the SLAM system on more real-world platforms.
Through establishing multi-camera manifold consistency constraints on motion trajectories, we effectively calculate metric-scale poses  during initialization and continuously refine metric-scale in the backend optimization.
Additionally, we further boost the accuracy and efficiency of the pose estimation through the learning-based feature association frontend and the multi-camera loop closure.
Comparisons with the state-of-the-arts demonstrate the effectiveness and robustness of MCVO on state estimation. Future work will concentrate on deploying MCVO across more platforms and in diverse scenarios while simultaneously developing dense depth reconstruction using multi-cameras. 
 

\bibliographystyle{IEEEtran.bst}
\bibliography{references}

\end{document}